\documentclass{llncs}
\usepackage{makeidx}  
\usepackage{graphicx}
\usepackage[compatibility=false]{caption}
\usepackage{subcaption}

\begin{document}

\mainmatter

\title{Electricity Load Forecasting - An Evaluation of Simple 1D-CNN Network Structures}
\titlerunning{Electricity Load Forecasting}

\author{Christian Lang\inst{1,2} \and Florian Steinborn\inst{1} \and Oliver Steffens\inst{2} \and Elmar W. Lang\inst{1}}

\authorrunning{Christian Lang et al.}
\tocauthor{Christian Lang, Florian Steinborn, Oliver Steffens, Elmar W. Lang}
\institute{Regensburg Universit\"at, Regensburg, Germany,\\
\email{christian3.lang@ur.de},
\and
OTH Regensburg, Regensburg, Germany}

\maketitle

\begin{abstract}
This paper presents a convolutional neural network (CNN) which can be used for forecasting electricity load profiles 36 hours into the future. In contrast to well established CNN architectures, the input data is one-dimensional. A parameter scanning of network parameters is conducted in order to gain information about the influence of the kernel size, number of filters, and dense size. The results show that a good forecast quality can already be achieved with basic CNN architectures. The method works not only for smooth sum loads of many hundred consumers, but also for the load of apartment buildings. 
\end{abstract}

\section{Introduction}

There is no dispute in the scientific community that human-induced climate change is real. The effects of climate change are for example rising sea levels, an increasing CO$_2$ concentration in the atmosphere, and more regularly occurring extreme weather events, to name only few of them.\cite{ipcc:2018full,ipcc:2012full} To slow down and stop the global warming, it is crucial to reduce the generation of greenhouse gases, especially in energy production.
One of the keys to accomplish a reduction is to establish more renewable energies in the energy market. By doing so, power plants that produce high levels of CO$_2$, like coal power plants, can in the long term be substituted by renewable energy sources. Another key to minimise the emission of greenhouse gases is to decrease the total energy consumption and to increase energy efficiency in consumption and production.

\begin{figure}[h]
	\centering
	\includegraphics[width=0.6\textwidth]{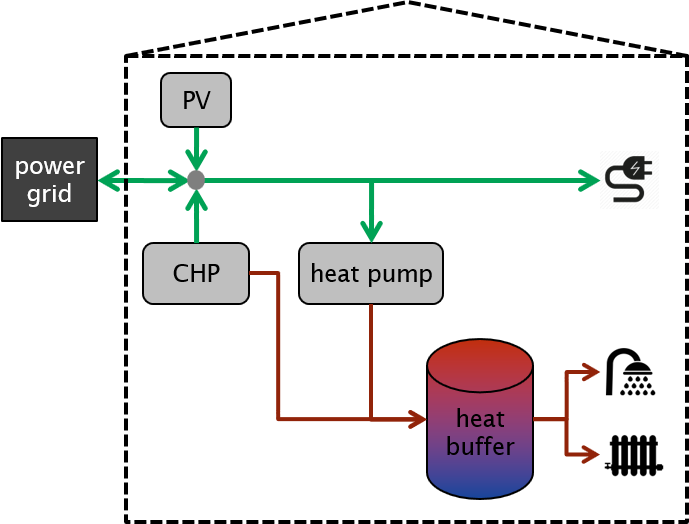}
		\caption{Schematic of the new heating system. The red lines symbolise heat transport using water and the green lines symbolise electricity transport.}
		\label{fig:heat_schematics}
\end{figure}

In the research project MAGGIE \cite{maggie:solaresbauen,maggie:enargus}, we try to address all of the above mentioned challenges. The goal of the research project is to energetically modernise existing historic apartment buildings and draft a concept for sector coupling in city districts. In the first step, one exemplary building will be modernised and evaluated. Afterwards, the whole city district will be modernised in a similar manner. In order to decrease the heat consumption of the building the thermal insulation is renewed and in the course of the research project new insulations are in development.
In addition, a new heating system (see fig.\,\ref{fig:heat_schematics}) with an innovative control system is implemented. This heating system allows increases in energy efficiency and can help integrate renewable energy sources into the power market. The core of the system is a combined heat and power plant (CHP), and a heat pump. All of them generate thermal energy, the heat pump from electricity and the CHP from fuel. The thermal energy is used to heat the water of a buffer storage, which is then, in combination with a heat exchanger, used as process and drinking water. In addition, the CHP generates electricity, as does a photovoltaic system (PV system) installed on the roof of the building, which can then be used to either power the heat pump or supply the habitants with electricity. A connection to the power grid receives surplus electricity and ensures there is always enough electricity available.

By utilising both forms of generated energy, the total energy efficiency of the system is nowadays higher than that of a conventional heating system, for example a gas-fired boiler, in combination with electricity from the power grid. \cite{buch:sterner}

All parts of the energy system are monitored continuously and can be controlled independently and remotely by the control system, which allows one to shift the production and consumption of heat and electricity in time and between the participants of the system by heating and using the water at the needed times. This allows for optimisation of the machine schedules depending on an optimisation target. Those targets can, for example, be self-sufficiency or cost-reduction.

After the modernisation of the entire city district, the energy systems of all houses in the district or even of several districts can act as a virtual power plant (VPP). This VPP can then work as a base load power plant or can help to stabilise the power grid and therefore assist to integrate renewable energies.

The main challenge of the system consists in knowing the electric and heat load of the building and its inhabitants. The loads are crucial for schedule optimisation as the feed-in into the power-grid and the consumption from the power grid have to be reported to the power grid operator the day before at midday in a 15 minute grid. Deviations in the heat load can be buffered with the heat buffer, deviations in the electric load, however, cannot be buffered in any way. Therefore, the focus of this paper is on forecasting electricity loads.

\section{Smart Meter Data}
\label{subsec:dataset}
In two directives \cite{eu.dir:1,eu.dir:2}, the EU outlined their decision to establish SmartMeter devices in the energy sector throughout the entire European Union with the aim to enable customers to better monitor and manage their consumptional behaviour. A SmartMeter, in contrast with a conventional electric meter, records the energy consumption at least every 15 minutes or in even shorter periods. In this paper, the data of the CER Smart Metering project \cite{cer.trial} is used. The dataset consists of individual SmartMeter data from over 5\,000 Irish homes and businesses recorded for 18 months.

\begin{figure}[h]
	\centering
	\begin{subfigure}{0.45\textwidth}
		\centering
		\includegraphics[width=\textwidth]{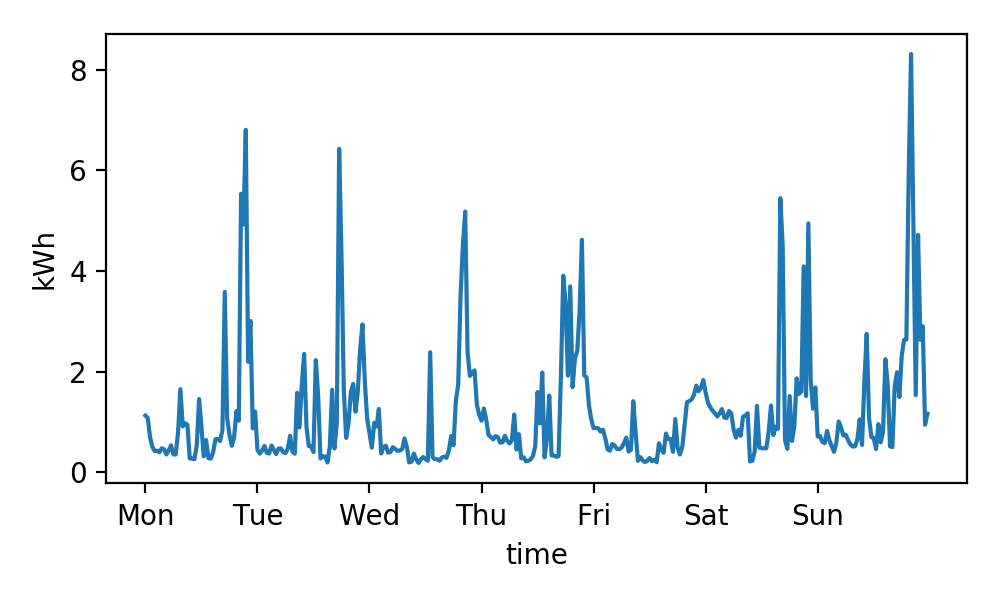}
		\caption{Load of a single household}
	\end{subfigure}
	\begin{subfigure}{0.45\textwidth}
		\centering
		\includegraphics[width=\textwidth]{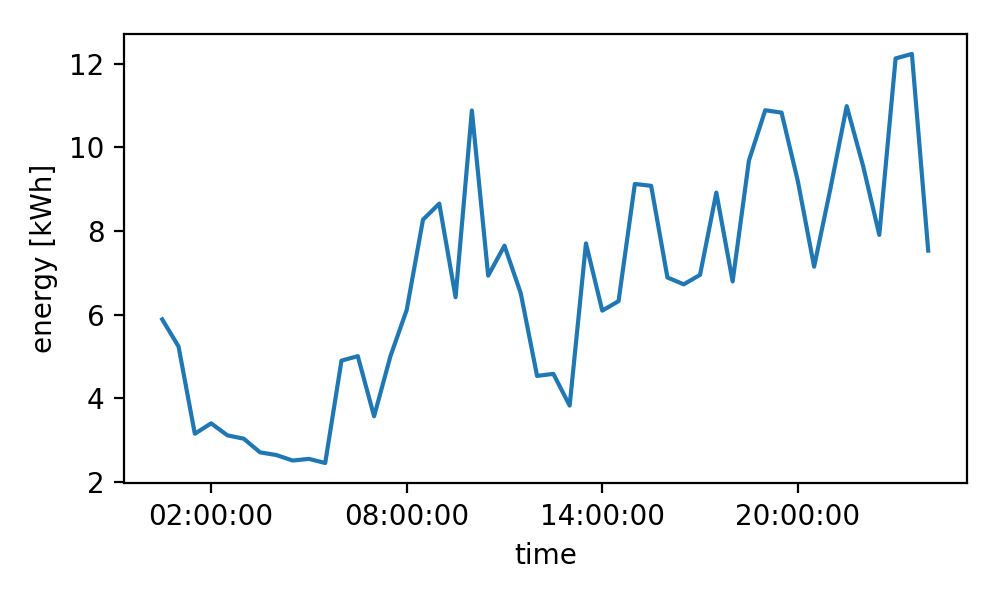}
		\caption{Combined load 15 households}
	\end{subfigure}
	\\
	\begin{subfigure}{0.45\textwidth}
		\centering
		\includegraphics[width=\textwidth]{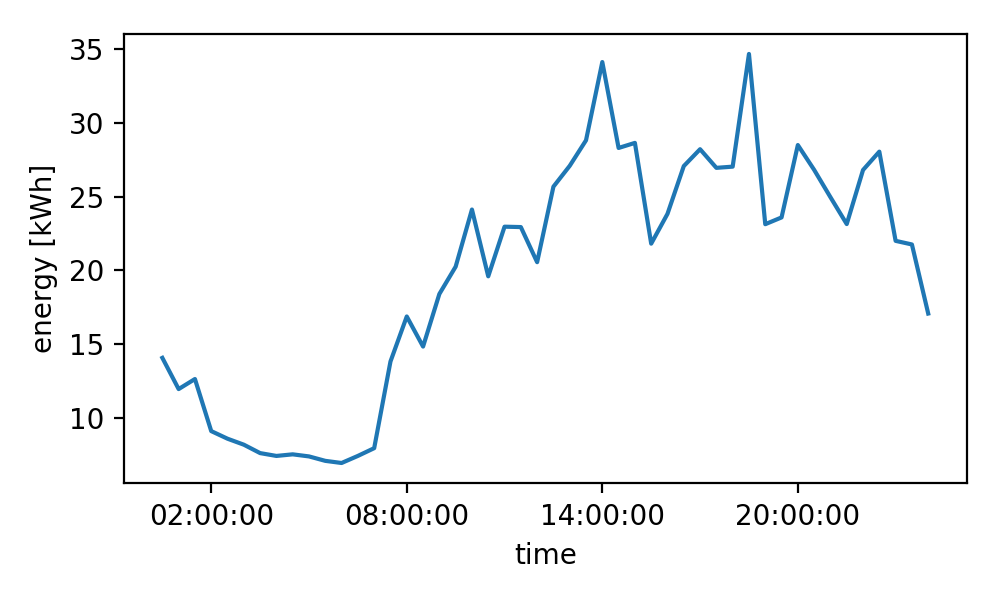}
		\caption{Combined load 40 households}
	\end{subfigure}
	\begin{subfigure}{0.45\textwidth}
		\centering
		\includegraphics[width=\textwidth]{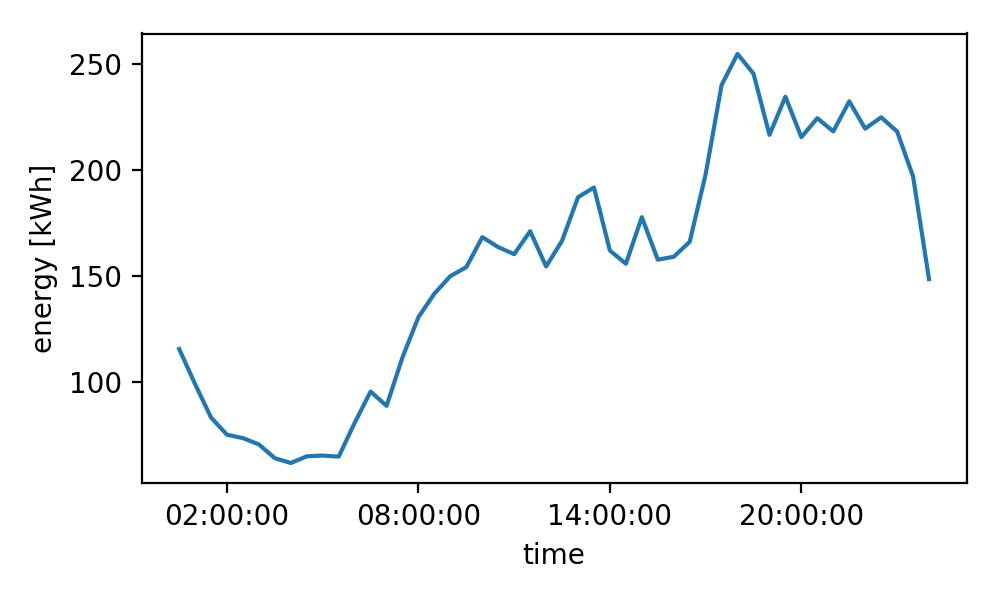}
		\caption{Combined load 350 households}
	\end{subfigure}
	\caption{(a) shows a load time series of a single household. (b)-(d) show each an exemplary day of the combined load time series. The different extracts display how volatile the load of a single household is and that the volatility decreases when households are combined.}
	\label{fig:loads}
\end{figure}

As the electric load of a single household is highly volatile and therefore impossible to predict, sum load time series of 15, 40, and 350 randomly picked households were created. Those time series correspond to a small apartment building, a big apartment building, and a whole city district. Figure \ref{fig:loads} shows an exemplary day of the mentioned time series. In this work we focus on the load time series of 40 and 350 households.

\section{Importance of Time Series Forecasting}
The electric load forecast is crucial in order to fully utilise the possibilities of the implemented heating system and similar systems. Without a good forecast, a part of the heat buffer capacity has to be withheld in order to balance the deviation in the electric load by the CHP. The prediction horizon is $h=144$ samples as $36\,$h have to be predicted in a $15\,$min grid.

There are already several publications about time series forecasting and short-term load forecasting (STLF).\,\cite{stlf1,stlf2} However, most of the methods predict either only one or very few time steps in the future, or are applied on load time series of whole cities/states which are, due to the properties of statistics, way smoother than the load time series of one building. Those smooth time series can be described properly with statistical methods when external factors (e.g. the weather) are taken into account. Therefore, their shape and features are also easy for neural networks to learn. None of the methods mentioned in recent publications, however, are designed to predict the electric load of only one building.

In the next chapter we propose the use of Convolutional Neural Networks (CNN) for time series prediction and report the first results of different network structures.

\section{Convolutional Neural Networks}

Convolutional Networks, in the way they are used today, were first introduced by LeCun et\,al.\cite{cnn.lecun} for zip code recognition. Since then, they were further developed and are now the standard for image and pattern recognition.

CNNs consist of convolutional layers, pooling layers, and fully-connected layers. In the convolutional layers, a set of feature maps, also called activation maps, are created. Each neuron in the feature map is only connected to a subset of neurons in its input layer. All neurons of the feature map share the same weights, thereby reducing the number of parameters significantly compared to a fully-connected neural network. In the most common CNN architectures, pooling layers alternate with convolutional layers. The pooling layer reduces the spatial dimension of the feature maps for the next computational steps in order to minimise the computational load and to avoid overfitting. At the end of the network, after an arbitrary number of the prior layers, fully-connected layers combine the resulting feature maps and return a classification measure. \cite{cnn.lecun,cnn.review}

\section{Forecasting with CNNs}

\begin{figure}
	\centering
	\includegraphics[width=0.5\textwidth]{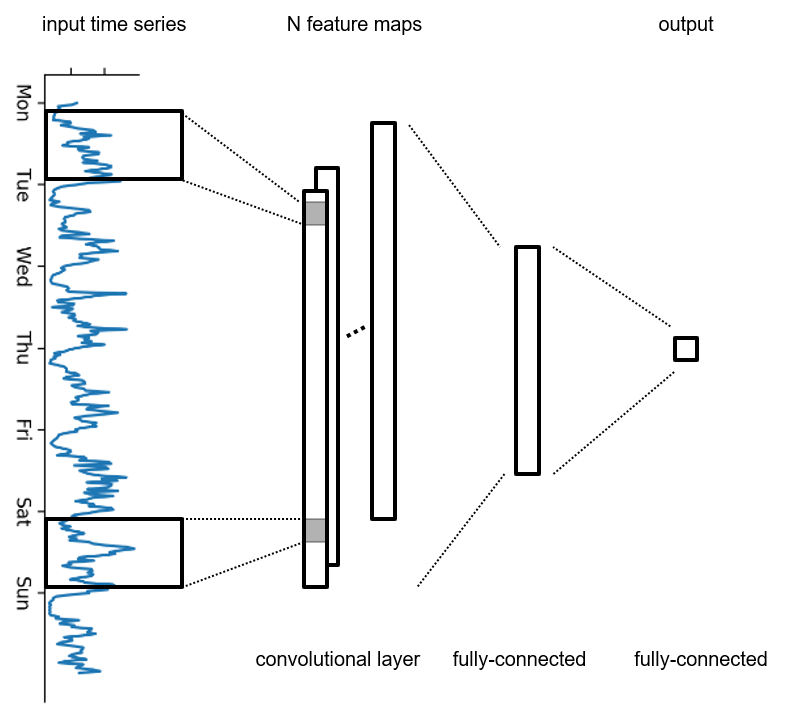}
	\caption{Principle architecture of the used neural network.}
	\label{fig:network}
\end{figure}

CNNs are traditionally used for image and pattern recognition by extracting features from two-dimensional data. In our research, we use a similar architecture for the forecasting of one-dimensional time series. The basic idea is that the convolutional layers extract features. These features are then combined by one or more fully-connected layers, and finally a forecast is created based on the classification of the last fully-connected layer (see fig\,\ref{fig:network}). The pooling layers are omitted because an excessive amount of parameters is not a problem for one-dimensional data and the necessity of pooling layers is questioned in recent research\,\cite{cnn.pool}.

A forecast can be created in two different ways, either directly or iteratively. A direct forecast means the network generates the desired forecast at once. Thus, the number of neurons in the output layer equal the prediction horizon $h$. When the forecast is generated iteratively, only one time step is predicted by the network. Then, the predicted point is appended to the input data and the first data point of the input is cut off, so that the new input has the required shape. Based on the new input, the next point is predicted. This procedure is repeated until $h$ data points are predicted.

\section{Evalution of different network structures and training parameters}

In order to get a better understanding of how the 1D-CNNs process data and how the network architecture influences the results, are the first tests conducted with very basic networks. They are built from one convolutional layer followed by one fully-connected layer which directly calculates the output. The evaluation of these networks yield the basic training parameters that are used throughout the rest of this work.

The best results were obtained with batch-size $b = 128$ and epochs $e = 40$. $Nadam$\,\cite{nadam} is used as optimiser and the mean squared error (MSE) as loss-function. When using bigger batch-sizes, unwanted jumps in the training loss were observed regularly, and when using smaller batch-sizes, overfitting occured early during training.

\begin{figure}[ht]
	\centering
	\begin{subfigure}{0.45\textwidth}
		\centering
		\includegraphics[width=\textwidth]{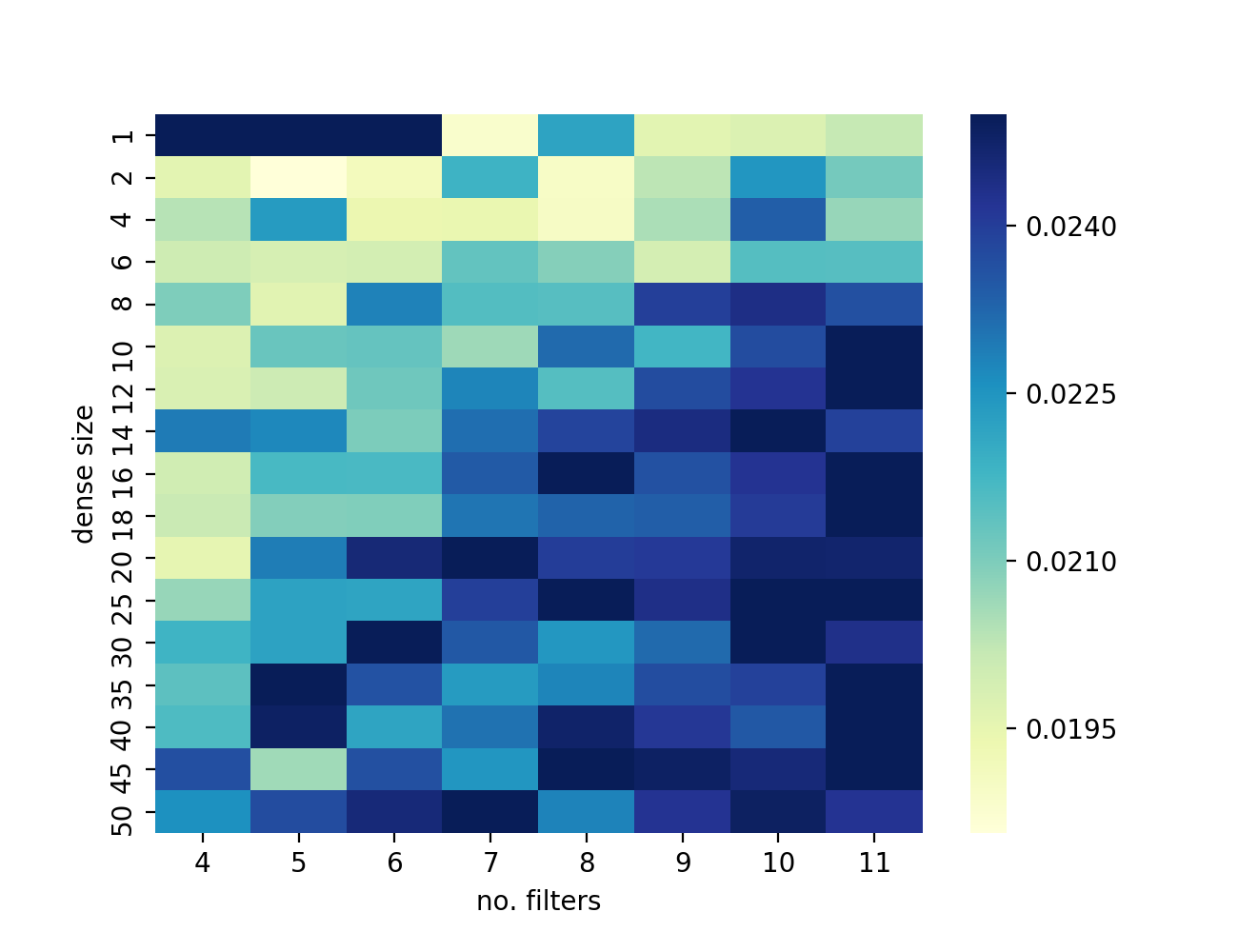}
		\caption{MSE dependent on the dense size and the number of filters.}
	\end{subfigure}
	\begin{subfigure}{0.45\textwidth}
		\centering
		\includegraphics[width=\textwidth]{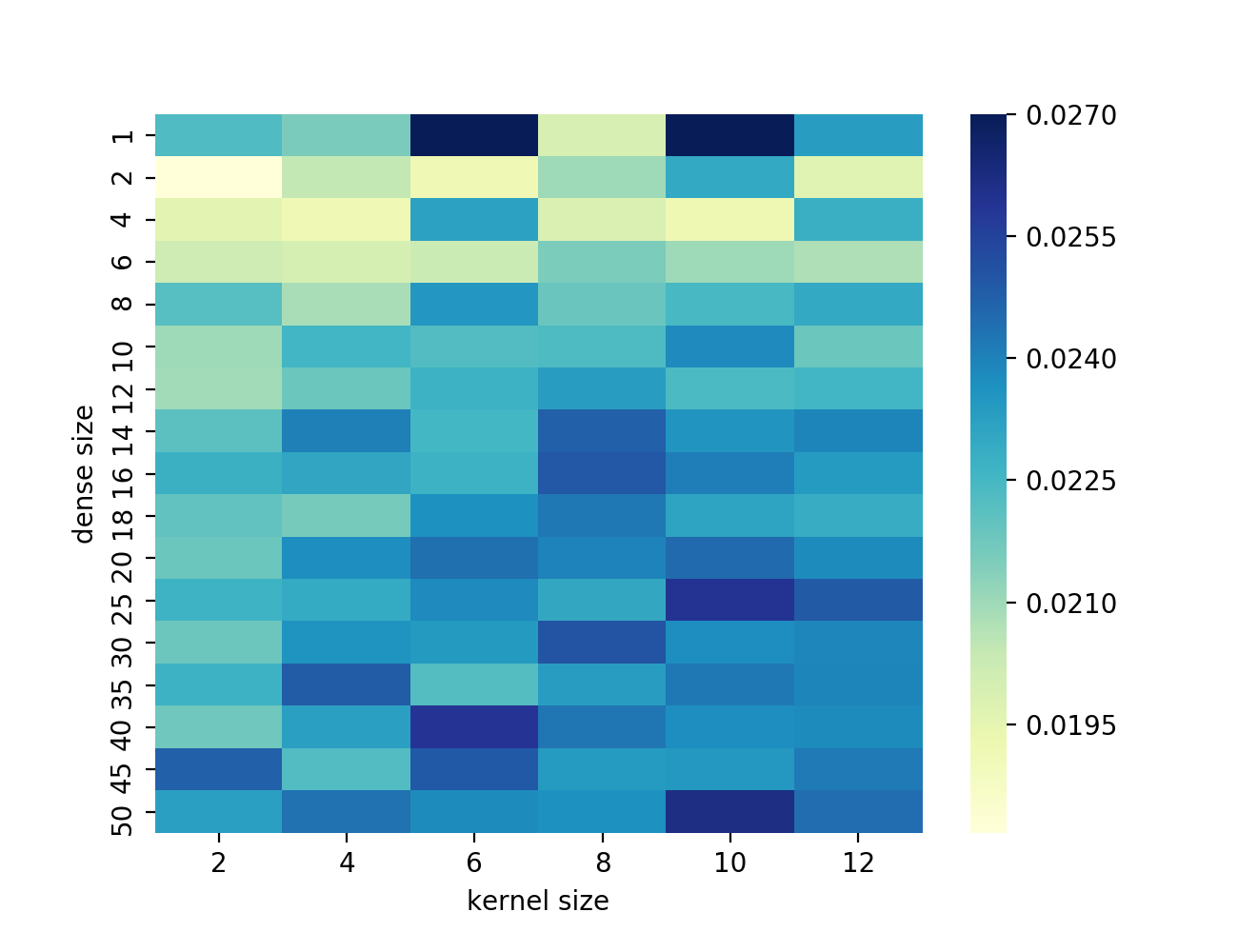}
		\caption{MSE dependent on the dense size and the kernel size}
	\end{subfigure}
	\begin{subfigure}{0.45\textwidth}
		\centering
		\includegraphics[width=\textwidth]{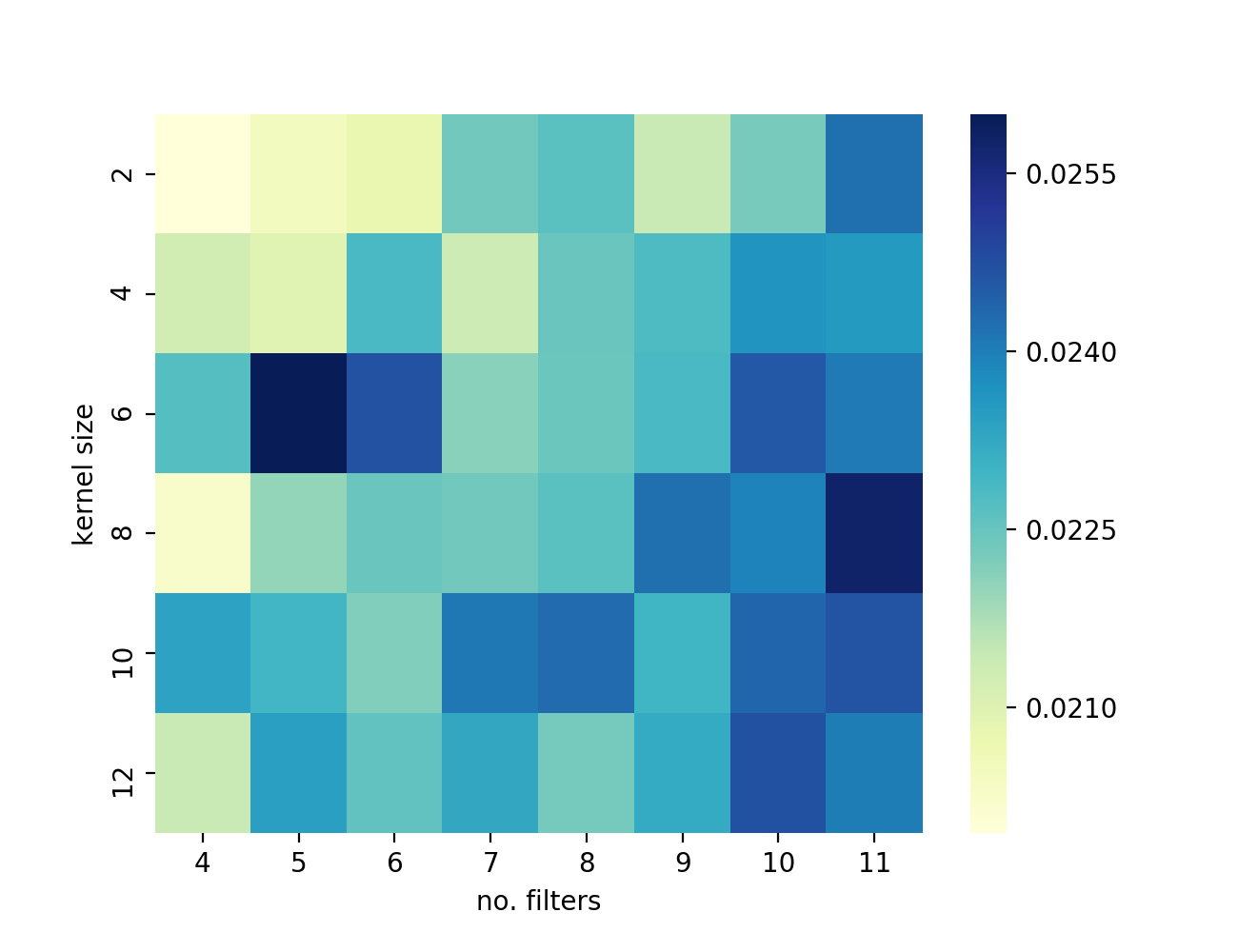}
		\caption{MSE dependent on the kernel size and the number of filter.}
	\end{subfigure}
	\begin{subfigure}{0.45\textwidth}
		\centering
		\includegraphics[width=\textwidth]{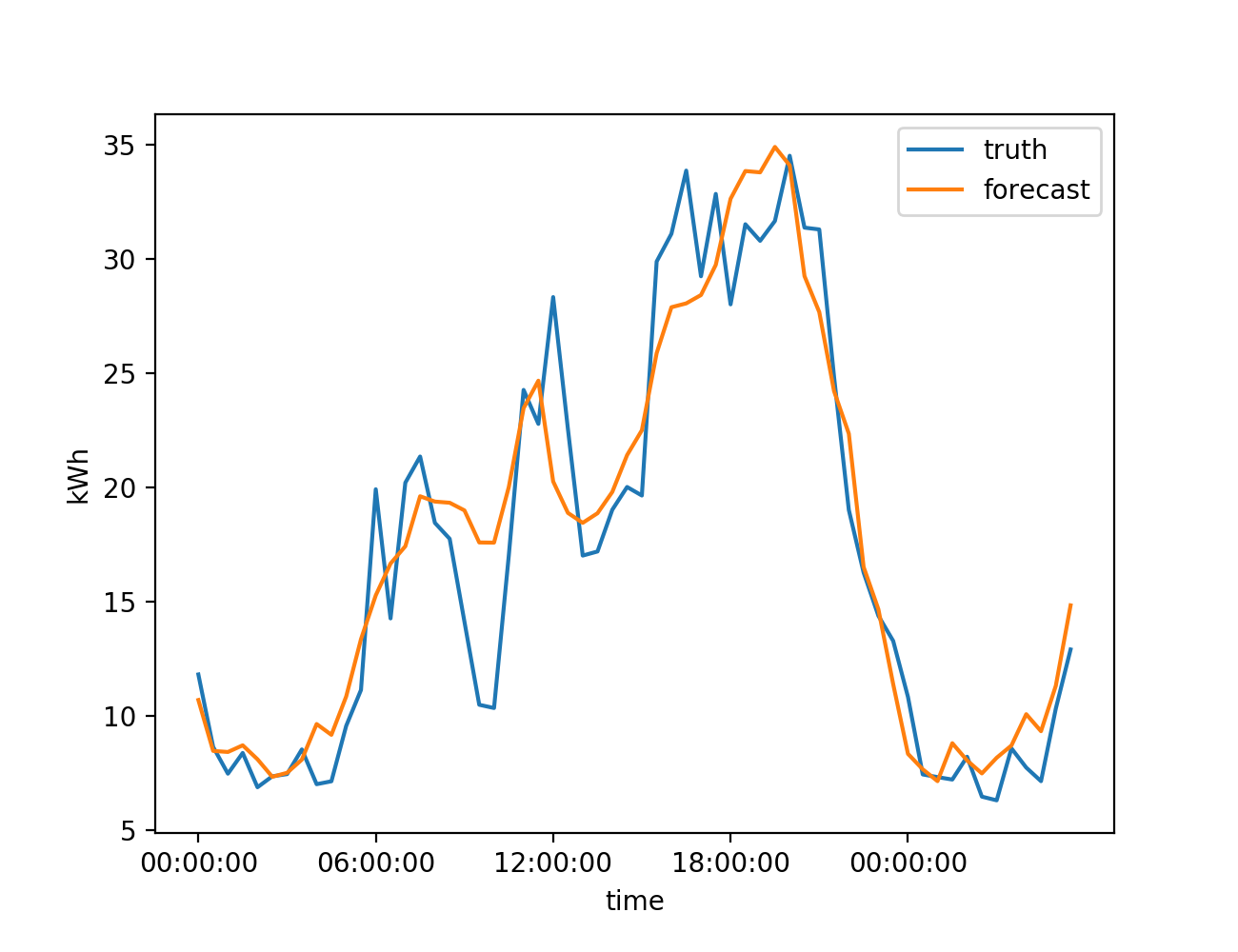}
		\caption{An exemplary forecast of the 350 households load on the validation data.}
	\end{subfigure}
	\caption{Heatmaps (a)-(c) show the MSE of the 40 household forecasts. The MSE values are a mean values across the third parameter. (d) shows a forecast using the trained CNN.}
	\label{fig:heatmaps40}
	
\end{figure}
In the next trials, an additional fully-connected layer was added in between the convolutional layer and the fully-connected output layer (see network architecture in fig. \ref{fig:network}). The additional fully-connected layer improved the forecast qualitiy independently from other network and training parameters. As this simple network already produces promising results, the convolutional layer is varied to further improve the forecasts.\\

The parameters that were varied are the kernel size and the number of filters. They describe how big the filters are, that sample over the time series, are and how many filters (each of them creates a feature map) are trained. The stride length is one. In addition, the influence of the first fully-connected layer is varied as well to identify how many features, from which the forecast is composed, exist.

\begin{figure}[ht]
	\centering
	\begin{subfigure}{0.45\textwidth}
		\centering
		\includegraphics[width=\textwidth]{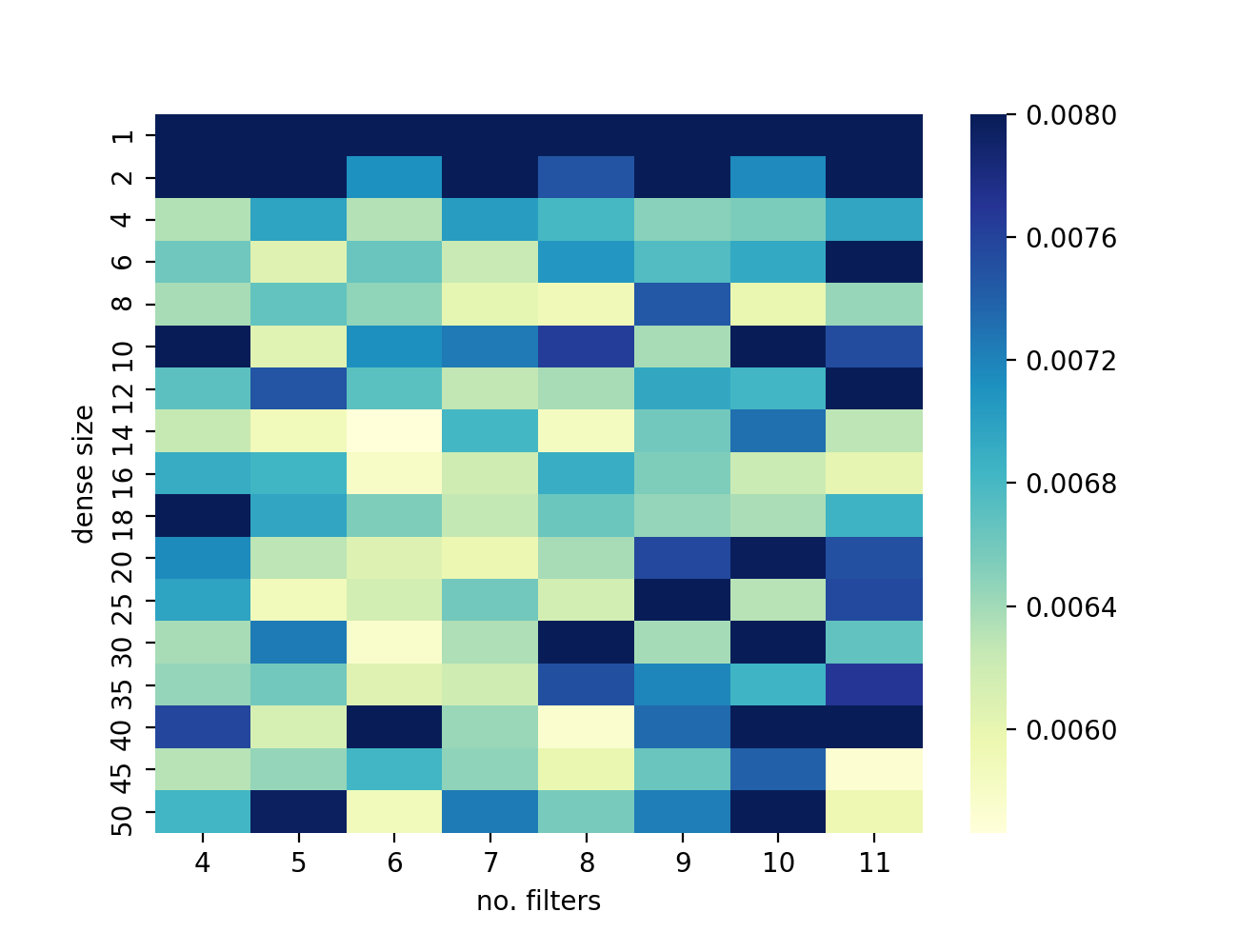}
		\caption{MSE dependent on the dense size and the number of filters.}
	\end{subfigure}
	\begin{subfigure}{0.45\textwidth}
		\centering
		\includegraphics[width=\textwidth]{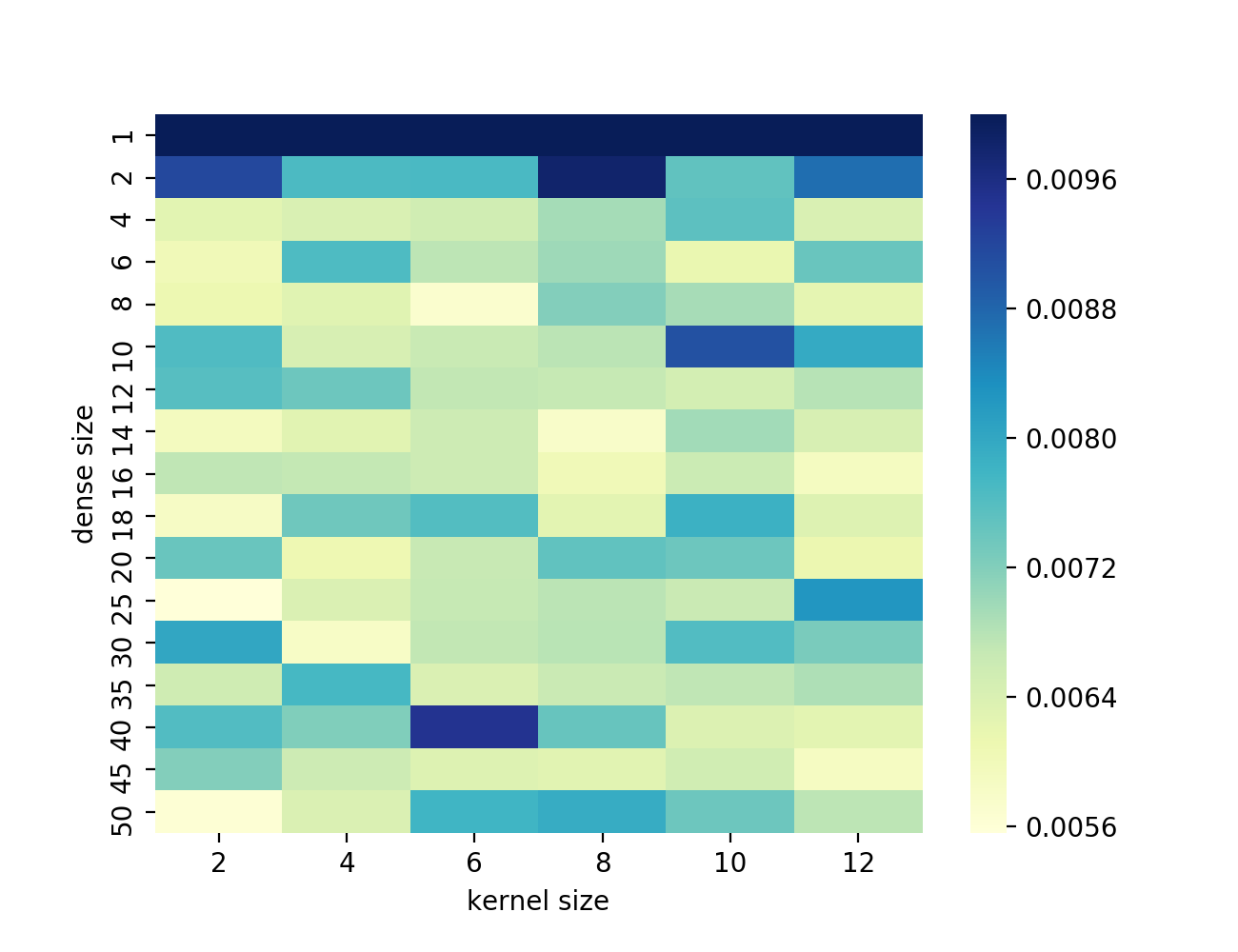}
		\caption{MSE dependent on the dense size and the kernel size}
	\end{subfigure}
	\begin{subfigure}{0.45\textwidth}
		\centering
		\includegraphics[width=\textwidth]{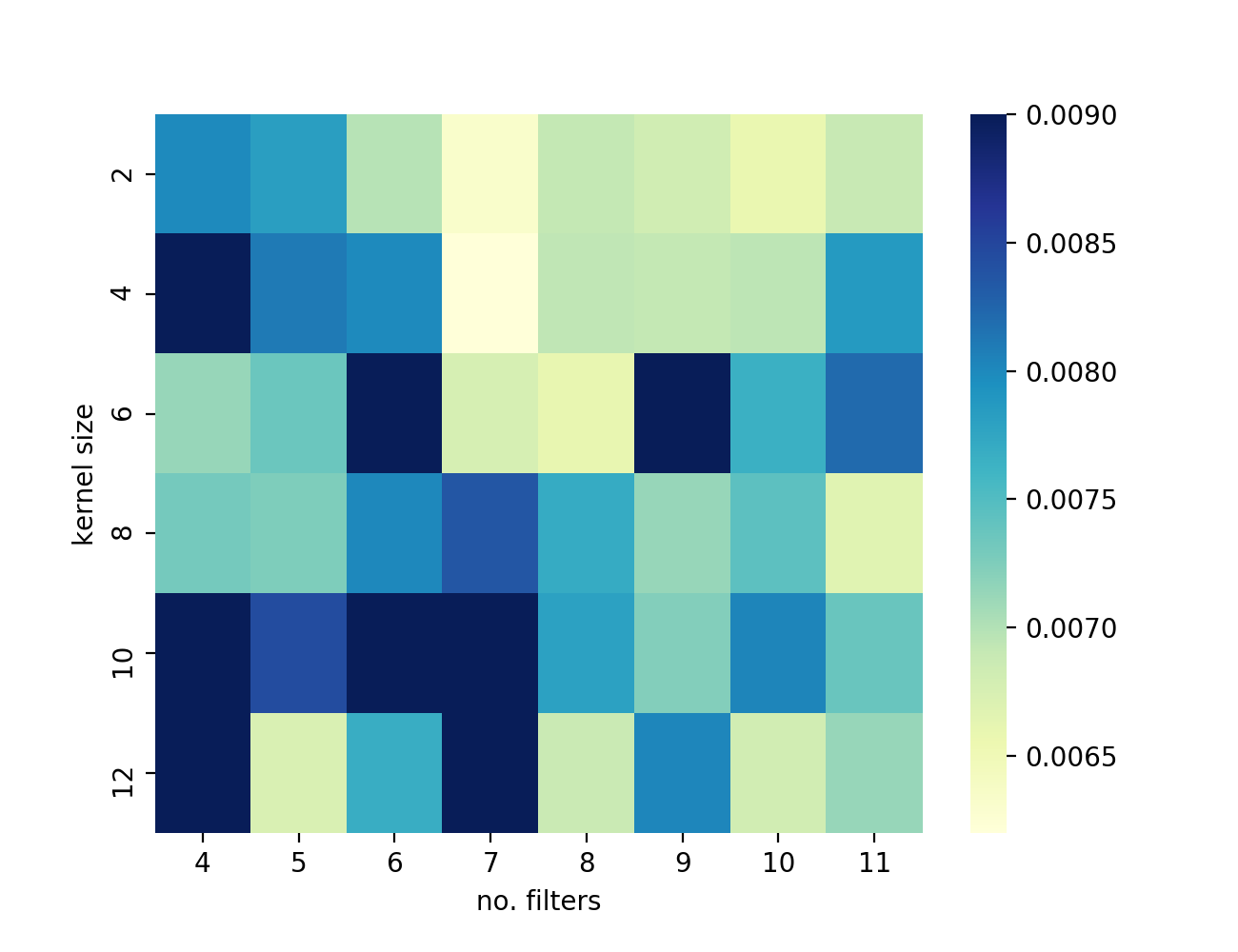}
		\caption{MSE dependent on the kernel size and the number of filter.}
	\end{subfigure}
	\begin{subfigure}{0.45\textwidth}
		\centering
		\includegraphics[width=\textwidth]{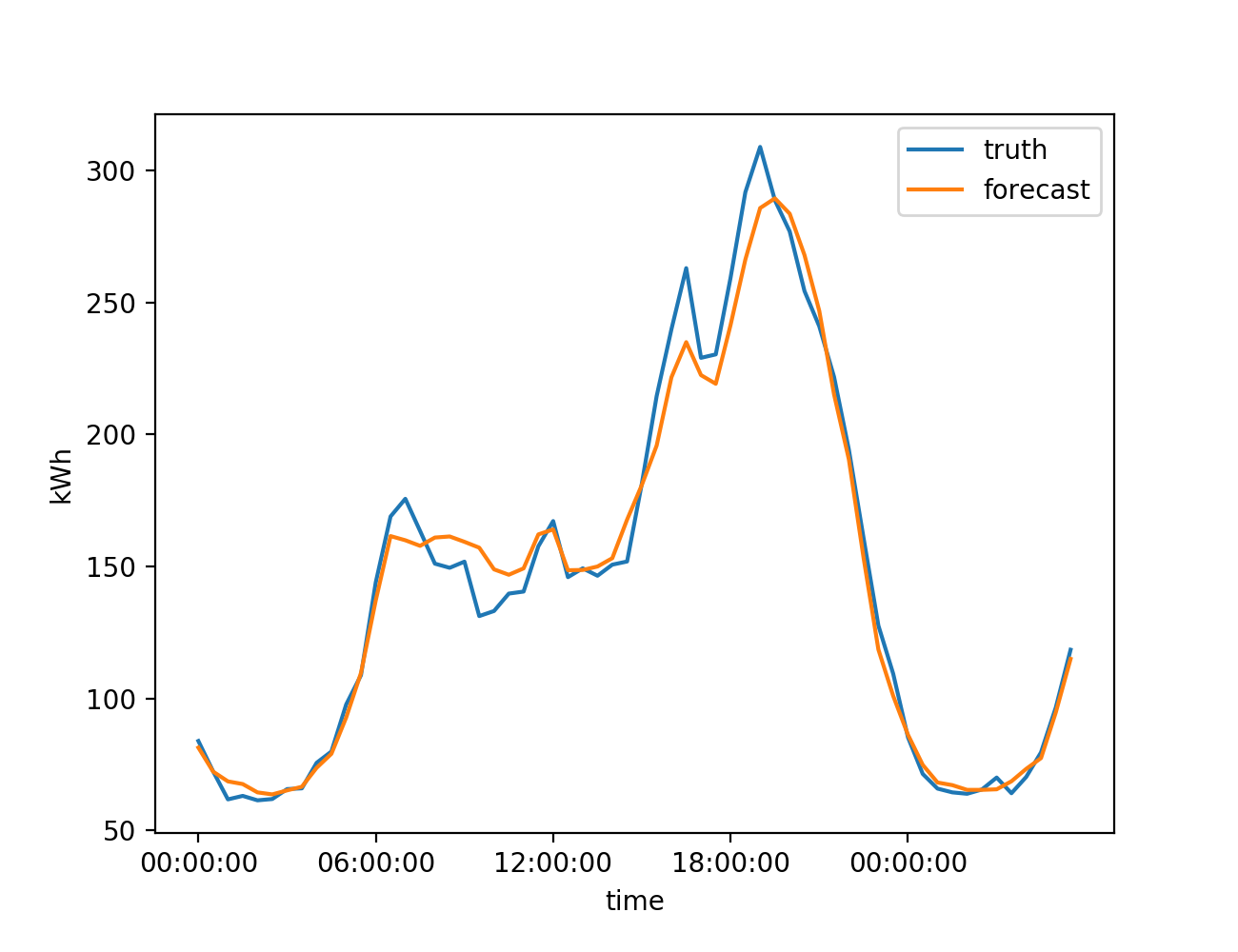}
		\caption{An exemplary forecast of the 350 households load on the validation data.}
	\end{subfigure}
	\caption{Heatmaps (a)-(c) show the MSE of the 350 household forecasts. The MSE values are a mean values across the third parameter. (d) shows a forecast using the trained CNN.}
	\label{fig:heatmaps350}
\end{figure}

\begin{figure}[ht]
	\centering
	\begin{subfigure}{0.32\textwidth}
		\centering
		\includegraphics[width=\textwidth]{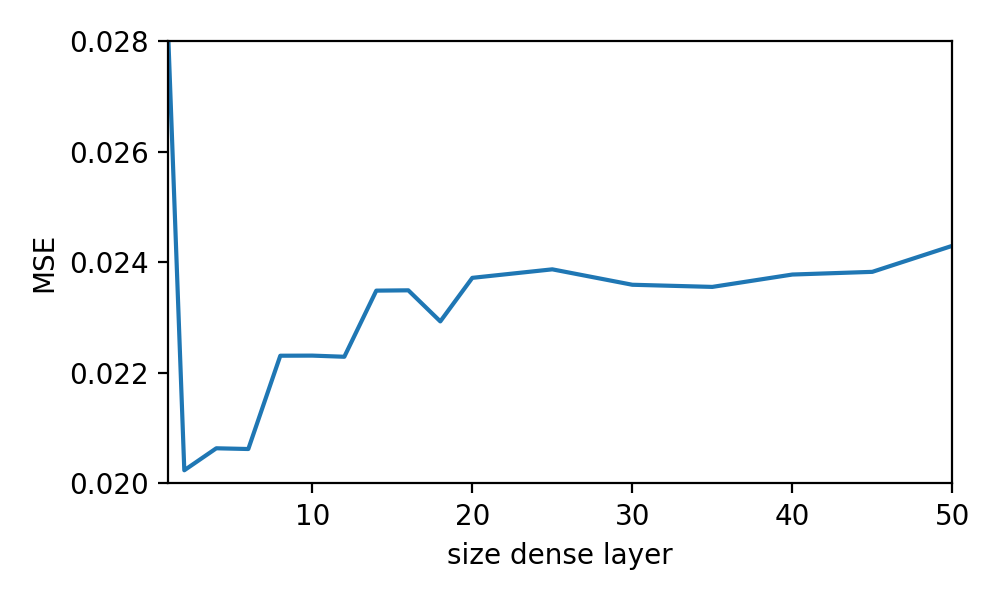}
		\caption{MSE relative to dense size; 40 households.}
	\end{subfigure}	\begin{subfigure}{0.32\textwidth}
		\centering
		\includegraphics[width=\textwidth]{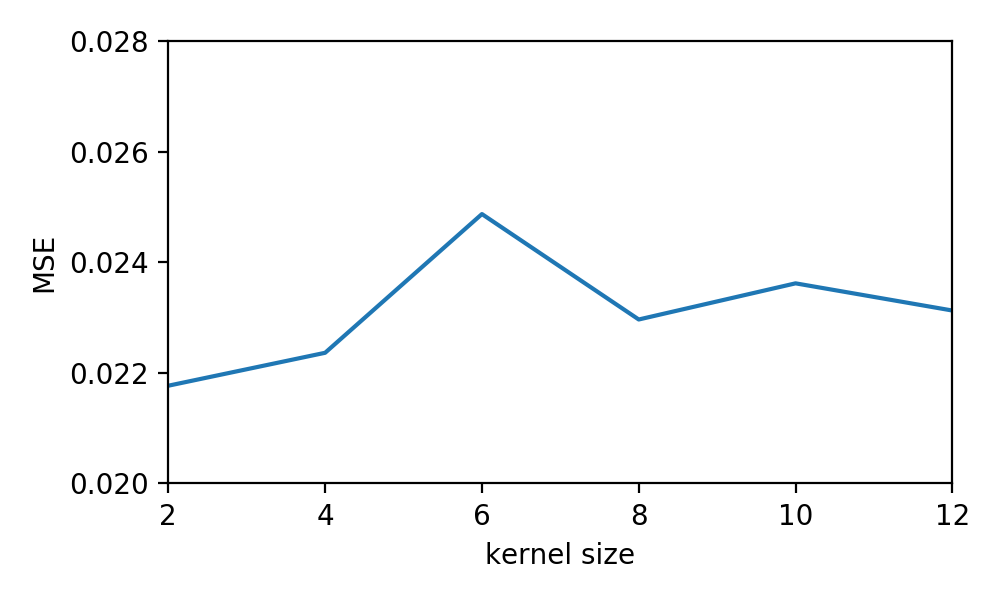}
		\caption{MSE relative to kernel size; 40 households.}
	\end{subfigure}
	\begin{subfigure}{0.32\textwidth}
		\centering
		\includegraphics[width=\textwidth]{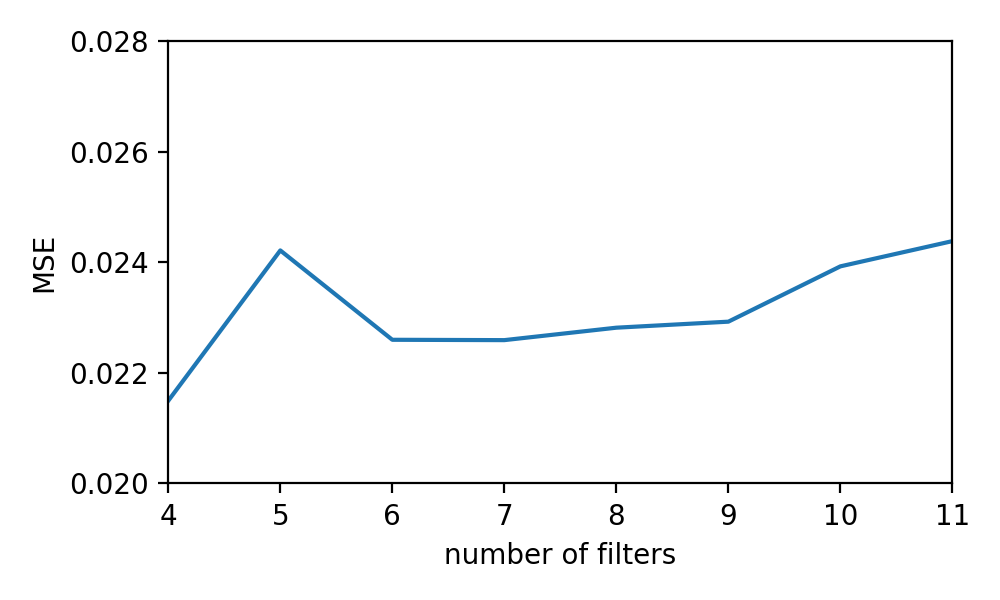}
		\caption{MSE realtive to number of filters; 40 households.}
	\end{subfigure}
	\\
	\begin{subfigure}{0.32\textwidth}
		\centering
		\includegraphics[width=\textwidth]{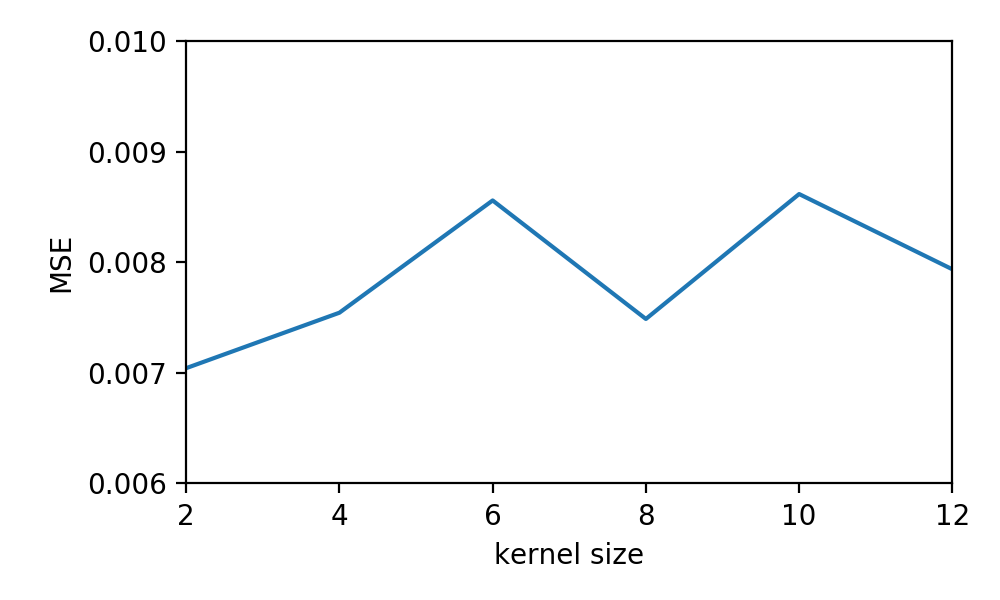}
		\caption{MSE relative to dense size; 350 households.}
	\end{subfigure}
	\begin{subfigure}{0.32\textwidth}
		\centering
		\includegraphics[width=\textwidth]{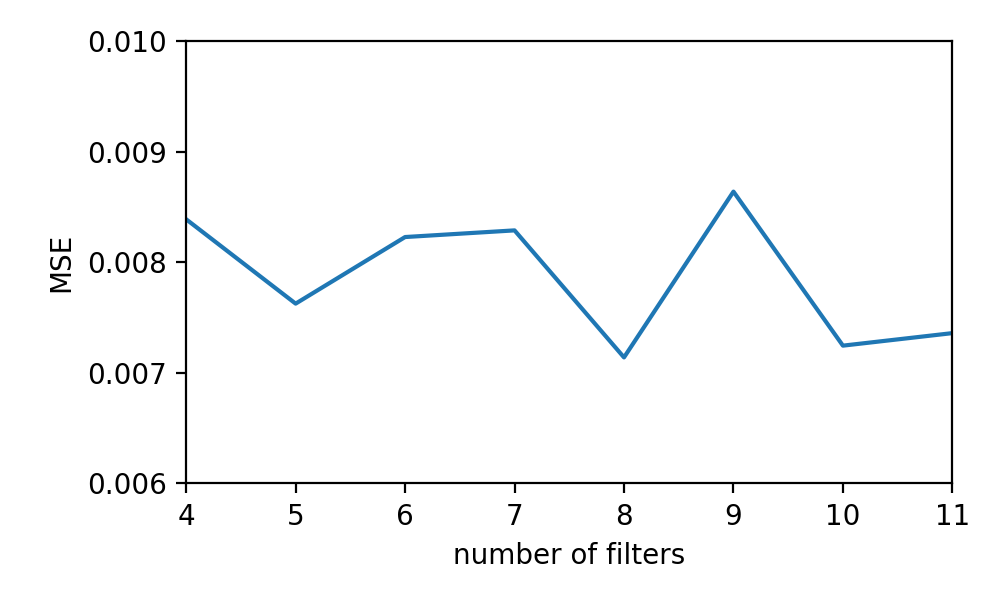}
		\caption{MSE relative to kernel size; 350 households.}
	\end{subfigure}
	\begin{subfigure}{0.32\textwidth}
		\centering
		\includegraphics[width=\textwidth]{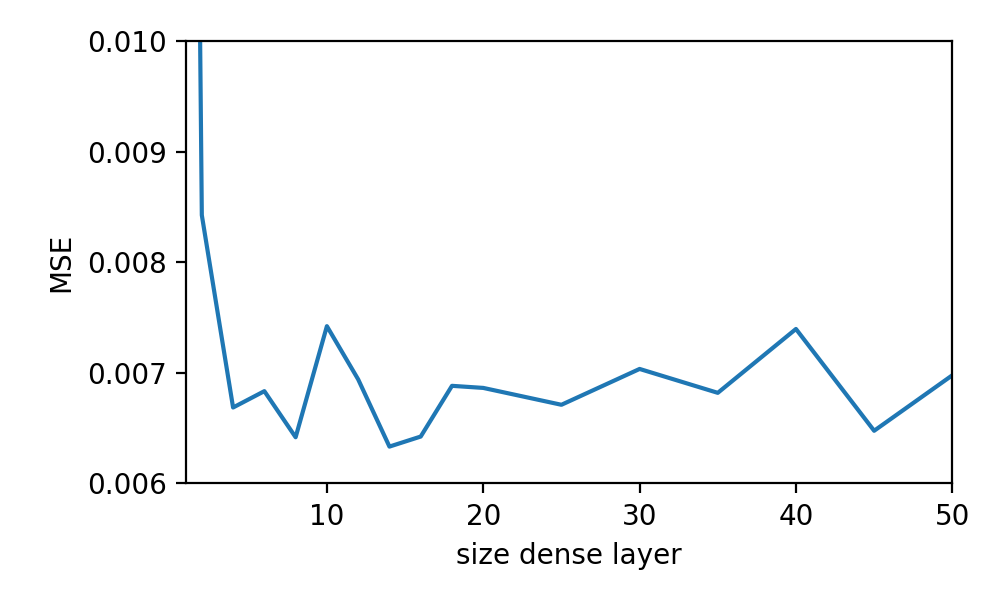}
		\caption{MSE relative to number of filters; 350 households.}
	\end{subfigure}
	\caption{The MSE averaged across the other two parameters for the 40 household load in (a)-(c) and for the 350 household load in (d)-(f).}
	\label{fig:mse}
\end{figure}

As is apparent from the heatmaps in figure \ref{fig:heatmaps40}, the four parameters have a crucial influence on the performance. On the heatmap plot (b), it can be seen that the best results can be achieved with a rather small fully-connected (also called dense) layer between the convolutional and the output layer. With an increasing dense size, the forecast results become unrelieable, probably overfitting occurs. In addition, the earlier conclusion that an additional fully-connected layer enhances the forecast quality is confirmed by the significantly worse performance of the network when $dense\_size = 1$. This basically equals a network with only one fully-connected layer. The heatmap (a) indicates that a large fully-connected layer compensates a small number of filters and vice versa. However, when both parameters that are chosen are too big, the MSE increases. Due to the high amount of trainable parameters in the network that come with a large dense size, it is advisable to use a small fully-connected layer with a larger number of filters, in order to minimise the computational load. There is no obvious conclusion regarding the kernel size. It seems that a kernel size which is too big or too small has a negative influence on the forecast quality. That impression is supported by the average MSE across dense size and number of filters (see fig. \ref{fig:mse}\,(a)-(c)).

The heatmaps of MSE of the 350 household load are illustrated in figure \ref{fig:heatmaps350}. The influence of the network parameters differs from the 40 household load. In addition to the confirmation that the additional fully-connected layer increases the forecast quality, it also becomes apparent that only with more than two neurons in the fully-connected layer good forecasts are possible. Furthermore, it seems that the number of filters and the kernel size only have a minor influence on the MSE. On heatmap (c), however, it appears that the most accurate forecasts are the ones with smaller kernel sizes. Figure \ref{fig:mse}\,(d)-(f) supports this assumption.

\section{Conclusion}

The network parameters dense size, number of filters, and kernel size were varied in a wide range. It can be conclude that the right set of parameters depends on the type of time series that is to be predicted.

The 350 household load time series can be forecasted properly with a CNN (see fig. \ref{fig:heatmaps350}\,(d)). When the size of the fully-connected layer chosen is larger than two, the network is quite robust against changes in the number of filters and kernel size.

A relieable forecast of the load time series of 40 households is possible with a rather simple CNN when the parameters are chosen correctly (see fig. \ref{fig:heatmaps40}\,(d)). It appears the time series can be described properly with 4 to 6 features as the best results were obtained with $dense\_size = 4...6$. Furthermore, it became apparent that with too many training parameters the forecast quality decreases, probably due to overfitting. 

The forecasters for both time series can already outperform the standard load profile, even though the network architecture is quite simple and no external factors have been taken into account yet. For volatile load profiles, simplicity in the network architecture seems to be the key for good forecasting results.

\section{Acknowledgments}

We thank the Federal Ministry for Economic Affairs and Energy (BMWi) for funding the project MAGGIE.

%

\clearpage

\begin{thebibliography}{5}
%
\bibitem {ipcc:2018full}
V. Masson-Delmotte, P. Zhai, H. O. P\"ortner, D. Roberts, J. Skea, P.R. Shukla, A. Pirani, W. Moufouma-Okia, C. Péan, R. Pidcock, S. Connors, J. B. R. Matthews, Y. Chen, X. Zhou, M. I. Gomis, E. Lonnoy, T. Maycock, M. Tignor, T. Waterfield (eds.): IPCC, 2018: Global warming of 1.5$^{\circ}$C. An IPCC  Special Report on the impacts of global warming of 1.5$^{\circ}$C above pre-industrial levels and related global greenhouse gas emission pathways, in the context of strengthening the global response to the threat of climate change, sustainable development, and efforts to eradicate poverty

\bibitem {ipcc:2012full}
Field, C.B., V. Barros, T.F. Stocker, D. Qin, D.J. Dokken, K.L. Ebi, M.D. Mastrandrea, K.J. Mach, G.-K. Plattner, S.K. Allen, M. Tignor, and P.M. Midgley (eds.): IPPC, 2012: Managing the Risks of Extreme Events and Disasters to Advance Climate Change Adaptation

\bibitem {maggie:solaresbauen}
Bundesministerium f\"ur Wirtschaft und Energie, Energiewende bauen, Solares Bauen: MAGGIE. https://projektinfos.energiewendebauen.de/projekt/energetisch-modernisieren-mit-solaraktiven-baustoffen-und-hybridem-heizsystem/

\bibitem {maggie:enargus}
Projekttr\"ager J\"ullich, EnArgus, Solares Bauen: MAGGIE.https://www.enargus.de\newline /pub/bscw.cgi/?op=enargus.eps2\&q=\%2201180590/1\%22\&v=10\&id=539378

\bibitem{buch:sterner}
M. Sterner, I. Stadler: Energiespeicher - Bedarf, Technologien, Integration. 2014 Springer Berlin Heidelberg.

\bibitem{eu.dir:1}
Directive 2006/32/EC of the European Parliament and of the Council of 5 April 2006 on energy end-use efficiency and energy services and repealing Council Directive 93/76/EEC. https://eur-lex.europa.eu/legal-content/EN/TXT/HTML/?uri=CELEX:32006L0032

\bibitem{eu.dir:2}
Directive 2009/72/EC of the European Parliament and of the Council of 13 July 2009 concerning common rules for the internal market in electricity and repealing Directive 2003/54/EC. https://eur-lex.europa.eu/legal-content/EN/TXT/HTML/?uri=CELEX:32009L0072

\bibitem{cer.trial}
Commission for Energy Regulation (CER). (2012). CER Smart Metering Project - Electricity Customer Behaviour Trial, 2009-2010 [dataset]. 1st Edition. Irish Social Science Data Archive. SN: 0012-00. www.ucd.ie/issda/CER-electricity

\bibitem{stlf1}
A. K. Srivastava, A. S. Pandey, D. Singh: Short-term load forecasting methods: A review. 2016 International Conference on Emerging Trends in Electrical Electronics \& Sustainable Energy Systems (ICETEESES), Sultanpur, 2016, pp. 130-138.
doi: 10.1109/ICETEESES.2016.7581373

\bibitem{stlf2}
B. Hayes, J. Gruber,M. Prodanovic: Short-Term Load Forecasting at the local level using smart meter data. 2015 IEEE Eindhoven PowerTech, Eindhoven, 2015, pp. 1-6.
doi: 10.1109/PTC.2015.7232358

\bibitem{cnn.lecun}
Y LeCun, B. Boser, J.S. Denker, D. Henderson, R.E. Howard, W. Hubbard, L.D. Jackel: Backpropagation Applied to Handwritten Zip Code Recognition. Neural Computation, 1(4):541-551, Winter 1989.

\bibitem{cnn.review}
N. Aloysius, M. Geetha: A review on deep convolutional neural networks. 2017 International Conference on Communication and Signal Processing (ICCSP), Chennai, 2017, pp. 0588-0592. doi: 10.1109/ICCSP.2017.8286426

\bibitem{cnn.pool}
J.T. Springenberg, A. Dosovitskiy, T. Brox, M. Riedmiller: Striving for Simplicity: The All Convolutional Net. ICLR (workshop track), 2015.

\bibitem{nadam}
T. Dozat: Incorporating Nesterov Momentum into Adam. 2016 ICLR 2016 workshop submission

\end{thebibliography}
\end{document}